%% file: main.tex
\documentclass[runningheads]{llncs}

\usepackage{eccv}

\usepackage{eccvabbrv}

\usepackage{graphicx}
\usepackage{booktabs}
\usepackage{hyperref}
\usepackage{url}
\usepackage{float}
\usepackage{multirow}
\usepackage{multicol}
\usepackage{amsmath}
\usepackage{amssymb}
\usepackage{wrapfig}

\usepackage[accsupp]{axessibility}  % Improves PDF readability for those with disabilities.

\newcommand{\sysname}{{\tt AdaDiff}}
\usepackage{hyperref}

\usepackage{orcidlink}

\begin{document}

\title{DeeDiff: Dynamic Uncertainty-Aware Early Exiting for Accelerating Diffusion Model Generation} 

\title{AdaDiff: Accelerating Diffusion Models through Step-Wise Adaptive Computation}
\titlerunning{AdaDiff}

\author{Shengkun Tang\inst{1} \and
Yaqing Wang\inst{2}\orcidlink{0000-0002-1548-0727} \and
Caiwen Ding\inst{3}\orcidlink{0000-0003-0891-1231}\and
Yi Liang\inst{2}\and
Yao Li\inst{4}\orcidlink{0000-0002-7195-5774}\and
Dongkuan Xu\inst{5}\orcidlink{0000-0002-1456-9658}
}
\authorrunning{Tang et al.}

\institute{MBZUAI\and Google  \and University of Minnesota, Twin Cities \and
UNC at Chapel Hill \and
North Carolina State University
}

\maketitle

\input{section/abstract}
\input{section/Intro}

\input{section/related}
\input{section/methods}

\input{section/experiments}
\input{section/conclusion}

\clearpage
\bibliographystyle{splncs04}
\bibliography{main}
\end{document}

%% file: section/abstract.tex
\begin{abstract}
Diffusion models achieve great success in generating diverse and high-fidelity images, yet their widespread application, especially in real-time scenarios, is hampered by their inherently slow generation speed. The slow generation stems from the necessity of multi-step network inference. While some certain predictions benefit from the full computation of the model in each sampling iteration, not every iteration requires the same amount of computation, potentially leading to inefficient computation. Unlike typical adaptive computation challenges that deal with single-step generation problems, diffusion processes with a multi-step generation need to dynamically adjust their computational resource allocation based on the ongoing assessment of each step's importance to the final image output, presenting a unique set of challenges.  In this work, we propose {\sysname}, an adaptive framework that dynamically allocates computation resources in each sampling step to improve the generation efficiency of diffusion models. To assess the effects of changes in computational effort on image quality, we present a timestep-aware uncertainty estimation module (UEM). Integrated at each intermediate layer, the UEM evaluates the predictive uncertainty. This uncertainty measurement serves as an indicator for determining whether to terminate the inference process. Additionally, we introduce an uncertainty-aware layer-wise loss aimed at bridging the performance gap between full models and their adaptive counterparts. Comprehensive experiments including class-conditional, unconditional, and text-guided image generation across multiple datasets demonstrate superior performance and efficiency of {\sysname} relative to current early exiting techniques in diffusion models. 
Notably, we observe enhanced performance on FID, with an acceleration ratio reduction of around 45\%. Another exciting observation is that adaptive computation can synergize with other efficiency-enhancing methods such as reducing sampling steps to accelerate inference. 
% Full code is released for reproduction.
% \footnote{Project Repo: \url{https://anonymous.4open.science/r/AdaDiff-DDB8}} 
\end{abstract}

%% file: section/Intro.tex
\section{Introduction}
% Diffusion models have shown significant performance improvement in generating diverse and high-fidelity images. A line of recent works~\cite{ho2020denoising,balaji2022ediffi,song2020ddim, kim2021soft, li2022diffusion, lu2022maximum, sehwag2022generating, austin2021structured} demonstrate the superiority compared with state-of-the-art GAN~\cite{brock2018large,lee2021vitgan, goodfellow2020gan} models on different tasks such as unconditional image generation~\cite{ho2020denoising}, image editing~\cite{li2022efficientdiff}, video generation~\cite{ho2022video} and text-guided image generation~\cite{ramesh2022hierarchical}. However, one drawback of diffusion models is the large computation requirement for progressive generation, which hinders the wide application of diffusion models. Specifically, the generation process of the diffusion model requires multi-step denoise where each step represents a full inference of the neural network. Such a procedure leads to a low speed of diffusion models compared to GAN models.
Diffusion models have shown significant performance improvement in generating diverse and high-fidelity images. A line of recent works~\cite{ho2020denoising,balaji2022ediffi,song2020ddim,kim2021soft,li2022diffusion,lu2022maximum,sehwag2022generating,austin2021structured} demonstrate their effectiveness on various tasks such as unconditional image generation~\cite{ho2020denoising}, image editing~\cite{li2022efficientdiff}, video generation~\cite{ho2022video}, and text-guided image generation~\cite{ramesh2022hierarchical}. However, one bottleneck of diffusion models in terms of applicability is their computational requirement for progressive generation. The generation process of a diffusion model involves multi-step denoising, where each step represents a full inference of the neural network. This multi-step denoising procedure results in slower generation speed, higher computational cost, and longer generation time compared to some other generative approaches, potentially impacting their applicability in scenarios requiring fast image generation.

% The speed of image generation within diffusion models is primarily determined by two factors: the number of generation steps and the one-step speed of model inference. In attempts to enhance generation efficiency, previous approaches~\cite{song2020ddim,lu2022dpm,lu2022dpm++} have predominantly focused on curtailing the number of sampling steps. Nonetheless, these methods still employ the same amount of network computation for each sampling step, therefore leading to suboptimal generation speed. Consequently, a natural question arises: \textit{is it necessary to use full models with all layers for all timesteps in diffusion generation?}

\begin{figure*}[t]
\begin{center}
\includegraphics[width=0.8\textwidth]{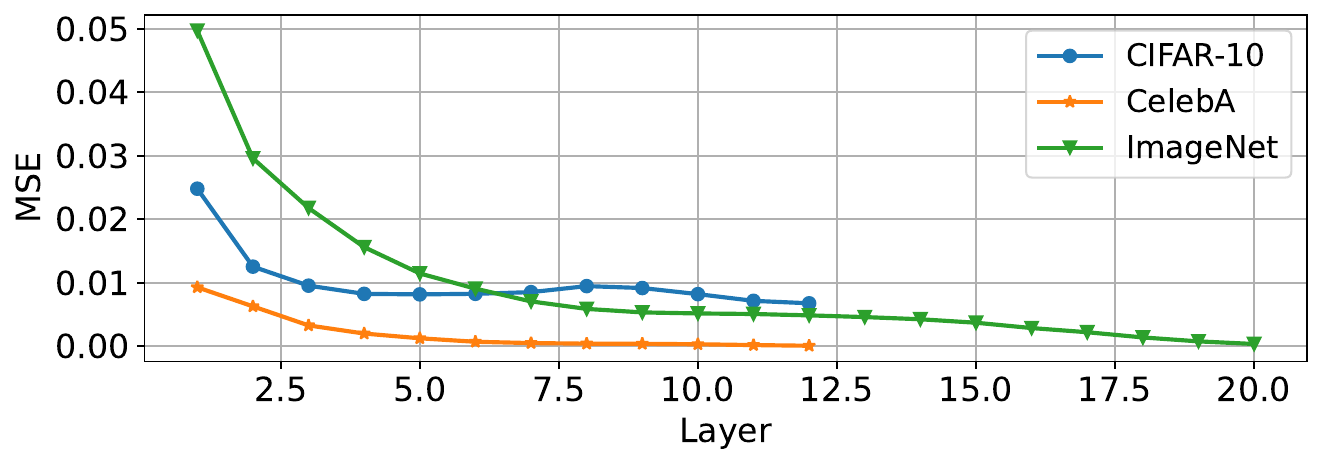}
\end{center}
\vspace{-0.3cm}
% \caption{Average MSE loss between intermediate layer outputs and final layer output across training samples in CIFAR-10, CelebA, and ImageNet datasets. The stable curve before the final layer shows shallow layers can potentially generate results comparable to the deep layers, motivating the possibility of adaptive computation approaches.}
\caption{Average Mean Squared Error (MSE) loss between intermediate layer outputs and the final layer output across testing samples in the CIFAR-10, CelebA, and ImageNet datasets. The relatively low MSE values before the final layer suggest that earlier layers have the potential to generate results comparable to those of the final layer. This observation motivates the exploration of adaptive computation approaches.}
\vspace{-0.4cm}
\label{Fig:inspiration}
\end{figure*}

The speed of image generation in diffusion models is primarily determined by two factors: the number of generation steps and the one-step inference speed. Previous approaches~\cite{song2020ddim,lu2022dpm,lu2022dpm++} have mainly focused on reducing the number of sampling steps to improve generation efficiency. However, these methods still use the same amount of network computation for each sampling step, which may not be optimal for achieving the best generation speed. This observation leads to an interesting question: \textit{is it necessary to use full models with all layers for all timesteps in diffusion generation?} Investigating this question could potentially lead to further improvements in the generation speed of diffusion models.

% In this paper, we demonstrate the existence of redundancy within the backbone networks in diffusion models.  The evidence is shown in Figure~\ref{Fig:inspiration}. The figure illustrates the trend of average Mean Square Loss (MSE) between each layer's intermediate output and the final layer's output across generation steps on CIFAR-10~\cite{krizhevsky2009learning}, CelebA~\cite{liu2015faceattributes}, and ImageNet~\cite{deng2009imagenet}. On CIFAR-10 and CelebA datasets, the backbone is the Transformer structure consists of 13 layers while on ImageNet, the backbone includes 21 layers.
% The loss drops sharply in the initial layers and then reaches a plateau in the middle layers across all datasets, indicating that the intermediate representations stabilize before the final layers. This suggests shallow layers can potentially generate results comparable to the deep layers, motivating the possibility of adaptive computation approaches.

In this paper, we explore the potential for optimizing the computational efficiency of diffusion models by investigating the redundancy within their backbone networks. Figure~\ref{Fig:inspiration} presents evidence supporting our hypothesis. The figure illustrates the trend of the average Mean Square Error (MSE) across all steps between the intermediate output of each layer and the final layer's output on CIFAR-10~\cite{krizhevsky2009learning}, CelebA~\cite{liu2015faceattributes}, and ImageNet~\cite{deng2009imagenet} datasets. For CIFAR-10 and CelebA, the backbone is a Transformer structure consisting of 13 layers, while for ImageNet, the backbone includes 21 layers. Interestingly, the MSE drops sharply in the initial layers and then plateaus in the middle layers across all datasets. This observation suggests that the intermediate representations stabilize before reaching the final layers, indicating that the shallow layers may have the capacity to generate comparable results as the deeper layers. This finding motivates us to explore the possibility of employing adaptive computation methods to improve the efficiency of diffusion models without compromising their generation quality.

Applying adaptive computation methods, such as early exiting, to diffusion models poses a significant challenge in determining the required computation at each step of the iterative generation process. Early exiting is an adaptive computation method that aims to save computation by early terminating the computation when the model has reached a sufficient level of confidence. However, diffusion models generate images by progressively denoising a latent representation over multiple timesteps, with each step conditioned on the previous one. Earlier steps focusing on coarse-grained structures may require less computation, but prematurely exiting at these steps can lead to error accumulation across subsequent steps, degrading the final image quality~\cite{hang2023efficient}. Conversely, exiting too late may result in unnecessary computations. To find the optimal balance between efficiency and quality, a mechanism is needed to assess the generated image's quality at each layer and determine the appropriate exit point, considering the multi-step generation process and the potential impact of error accumulation.

To address these challenges, we propose AdaDiff, a novel adaptive computation approach tailored to diffusion models. Unlike single-step computation allocation determining mechanisms in early exiting works, AdaDiff introduces a sampling uncertainty estimation module (UEM) to guide exits based on the sampling uncertainty of each layer at every timestep, taking into account the distinct stages of the sampling process. Furthermore, we introduce a new uncertainty-weighted, layer-wise loss that leverages the estimated uncertainties to reshape losses during training, bridging the performance divide between full models and their adaptive counterparts. Experiments demonstrate that AdaDiff reduces inference time with minimal performance degradation compared to state-of-the-art methods. Surprisingly, we find that the uncertainty-weighted loss can further boost performance when used separately in a full model setting, achieving better FID scores on CIFAR-10 and CelebA datasets.

Our main contributions are summarized as follows:
\begin{itemize}
% \item To the best of our knowledge, this is a pioneering work to extend early exiting to accelerate the inference of diffusion models. To this end, we propose a novel early exiting framework called AdaDiff with a valid assumption on different computation requirements of different sampling steps in diffusion models. It is worth noting that our AdaDiff framework can be easily plugged into any current both CNN-based and Transformer-based diffusion models. 

\item  To the best of our knowledge, this is a pioneering work that reshapes the landscape of accelerating diffusion model inference by introducing a novel adaptive computation approach. To this end, we propose AdaDiff, an innovative framework built upon the assumption that different sampling steps in diffusion models have varying computational requirements. What sets AdaDiff apart is its ability to be seamlessly integrated into any existing diffusion model, whether based on convolutional neural networks (CNNs) or Transformers, making it a versatile and powerful tool for enhancing the efficiency of diffusion models. Moreover, AdaDiff can be easily combined with efforts to reduce the number of sampling steps, further pushing the boundaries of acceleration in diffusion model inference.
    
% \item We introduce a timestep-aware uncertainty estimation module (UEM) in diffusion models to estimate the uncertainty of each prediction result from intermediate layers at each sampling step. The uncertainty estimated by UEM is utilized as an exiting signal to decide the timing of exiting. The results show that our proposed time-aware uncertainty estimation module is able to obtain accurate signals for exiting decisions.

\item In AdaDiff, we propose an effective way to determine the computation needed at each step during the progressive generation process in diffusion models. We introduce a timestep-aware uncertainty estimation module (UEM) that estimates the uncertainty of each prediction result from intermediate layers at each sampling step. The uncertainty estimated by the UEM is utilized as an exiting signal to decide the timing of exiting. Our results demonstrate that the proposed time-aware uncertainty estimation module successfully obtains accurate signals for exiting decisions, enabling AdaDiff to effectively balance speed and performance in diffusion model inference.
    
% \item To fill the performance gap between full models and early-exited models, we propose an uncertainty-aware layer-wise loss that reweights the layer-wise loss with estimated uncertainty from the uncertainty estimation module (UEM). Moreover, the experiments show that the loss strategy brings extra benefits to model training. Full-layer models trained with this loss achieve better performance than original models without early exiting.
    
\item Extensive experiments on unconditional,  class-conditional, and text-guided generation show that our method can largely reduce the inference time of diffusion models by at least 40\% with minimal performance drop on different datasets, demonstrating the effectiveness of our AdaDiff framework.
\end{itemize}

%% file: section/related.tex
\section{Related Work}
\noindent \textbf{Denoising Diffusion Models.} With the strong ability to generate high-fidelity images, diffusion model \cite{ho2020denoising, dhariwal2021diffusion, song2019generative, song2020score} achieved great success in many applications such as unconditional generation \cite{ho2020denoising, song2020score, karras2022elucidating}, text-guided generation \cite{balaji2022ediffi, rombach2022high, saharia2022photorealistic, ramesh2022hierarchical, chefer2023attend}, image inpainting \cite{lugmayr2022repaint} and so on.
Diffusion models are superior in modeling complex data distribution and stabilizing training in comparison with previous GAN-based models \cite{brock2018large, xu2018attngan, zhou2021lafite, zhu2019dm}, which suffer from unstable training. \cite{ho2020denoising} first proposed to utilize a neural network to estimate noise distribution. The backbone structure used in theirs and most other diffusion models is UNet \cite{ronneberger2015u}. More recently, a line of works \cite{bao2022all, peebles2022scalable} have also explored the application of  Transformers \cite{dosovitskiy2020image} as a backbone network, with U-ViT \cite{bao2022all} utilized the long skip
connection and leveraged Adaptive LayerNorm to achieve SoTA results on image generation.

\noindent \textbf{Efficiency in Diffusion.} One of the drawbacks of diffusion models is the low generation speed. On the one hand, diffusion models require multi-step (eg. 1000 steps) gradual sampling to generate high-fidelity images. On the other hand, the computation resource of one step is also expensive. Most existing works \cite{song2020ddim, luhman2021knowledge, salimans2022progressive, meng2022distillation, lu2022dpm, lu2022dpm++, lyu2022accelerating, san2021noise, watson2022fastsample} target on reducing sampling steps. Specifically, 
 \cite{luhman2021knowledge, salimans2022progressive, meng2022distillation} utilize the knowledge distillation method to insert multi-step teacher knowledge into the student model with fewer sampling steps. The drawback of these methods is they require large computation resources for distillation training. In contrast, 
 \cite{song2020ddim, lu2022dpm, lu2022dpm++, jolicoeur2021gotta, bao2022analytic} are able to reduce sampling steps to 50 with minimal performance loss, and no re-training is required. Besides on reducing sampling steps, \cite{li2023q} applies quantization to diffusion backbone to reduce GPU memory and increase backbone inference speed. \cite{lyu2022accelerating} applied early stop during training to accelerate diffusion.  However, all these methods utilize the full amount of network computation for each sampling step. \cite{moon2023early} proposed a static early exiting method to accelerate the inference of diffusion models. Still, such a predefined exiting strategy achieves suboptimal performance and efficiency. Instead, Our method dynamically allocates computation for inputs from different sampling steps, which accelerates the reverse denoising process adaptively.

\noindent \textbf{Early Exiting Strategy.} Early exiting~\cite{teerapittayanon2016branchynet} is a kind of method for neural network acceleration. The main assumption of early exiting is that different inputs require different computation resources. Existing early exiting methods~\cite{teerapittayanon2016branchynet, xin2020deebert, schuster2022confident, xin2021berxit, tang2022you} have achieved great success in improving efficiency. More concretely, 
 \cite{teerapittayanon2016branchynet} firstly appends a classifier after each convolutional neural network layer and utilizes entropy as a proxy to make exiting decisions. \cite{xin2020deebert} applies early exiting methods into BERT models and accelerates BERT with a slight performance drop. \cite{tang2022you} utilize early exiting strategies to accelerate vision language models and make early exiting decisions based on modalities and tasks. Moreover, \cite{xin2021berxit} proposes to learn to exit and extend their method into regression tasks. However, these models fail in diffusion models because of the time-series property of diffusion models in the denoising process. Our method proposes uncertainty-aware layer-wise loss to retain more information with fewer layers, achieving SoTA performance and efficiency compared with other methods.

%% file: section/methods.tex
\vspace{-0.4cm}
\section{Methods}
% \begin{figure*}[t]
% \begin{center}
% \includegraphics[width=1\textwidth]{fig/Arc_1.pdf}
% \end{center}
% \caption{The overview of our proposed DeeDiff. \textcolor{red}{[DK: add 2-3 sentences simply describing the flow via "first, second, ..."]}}
% \label{Fig:Architecture}
% \end{figure*}
\subsection{Preliminary}
Diffusion models generally include two processes: a forward noising process and a reverse denoising process. The forward process is a Gaussian transition, which gradually adds noise to the original image following a predefined noise scheduler. The final image is expected to follow standard Gaussian distribution. The reverse process 
progressively removes the noise from data sampled from standard Gaussian distribution to generate high-fidelity images. The backbone network is utilized to denoise from noised images.

Formally, let $\mathbf{x}_0$ be the sample from training data distribution $q(\mathbf{x}_0)$, namely $x_0 \sim q(\mathbf{x}_0)$. A forward process adds noise gradually to $\mathbf{x}_0$ for $T$ times, generating a series of intermediate noisy samples $\{\mathbf{x}_1, ..., \mathbf{x}_T\}$ following:
\begin{align}
    q(\mathbf{x}_t|\mathbf{x}_{t-1}) &= \mathcal{N}(\mathbf{x}_{t}; \sqrt{1\beta_t}\mathbf{x}_{t-1}, \beta_t\mathbf{I}), \ with \
    \mathbf{x}_t  = \alpha_t \mathbf{x}_0 + \beta_t \boldsymbol{\epsilon}.
\end{align}
%     q(\mathbf{x}_t|\mathbf{x}_{t-1}) = \mathcal{N}(\mathbf{x}_{t}; \sqrt{1\beta_t}\mathbf{x}_{t-1}, \beta_t\mathbf{I}) \\

%     \mathbf{x}_t  = \alpha_t \mathbf{x
% }_0 + \beta_t \boldsymbol{\epsilon}.

% \end{equation}
% \begin{equation}
%     \mathbf{x}_t  = \alpha_t \mathbf{x
% }_0 + \beta_t \boldsymbol{\epsilon}.
% \end{equation}
where $\beta_t \in (0,1)$ is the variance scheduler and controls the level of the Gaussian noise added to data in each step, $\boldsymbol{\epsilon}$ refers to the noise sampled from standard Gaussian distribution $\mathcal{N}(0, \mathbf{I})$. Each forward step only depends on the previous step. Therefore, the forward process follows the Markov Chain property. Moreover, as long as $T \rightarrow \infty$, $\mathbf{x}_T$ approaches an isotropic Gaussian distribution.

The reverse process is formulated as another Gaussian transition, which removes noise in noisy images and restores the original images. However, since the reserve conditional distribution $q(\mathbf{x}_{t-1}|\mathbf{x
}_t)$ is unknown at this point, diffusion models utilize neural networks to learn the conditional distribution $p_\theta(\mathbf{x}_{t-1} | \mathbf{x}_t)$:
\begin{equation}
    p_\theta(\mathbf{x}_{t-1} | \mathbf{x}_t) = \mathcal{N}(\mathbf{x}_{t-1}; \tilde{\boldsymbol{\mu}}_{\theta,t}(\mathbf{x}_t), \tilde{\boldsymbol{\Sigma}}_{\theta, t)}
\end{equation}
$\tilde{\boldsymbol{\mu}}_\theta$ and $\tilde{\boldsymbol{\Sigma}}_{\theta, t}(\mathbf{x}_t)$  are 
originally predicted statistics by backbone models. 
\cite{ho2020denoising} sets $\tilde{\Sigma}_{\theta, t}(\mathbf{x}_{t})$ to the constant $\tilde{\beta_t}^2\mathbf{I}$, and $\tilde{\boldsymbol{\mu}}_\theta$ can be formulated as the linear combination of $\mathbf{x}_t$ and backbone noise estimation model $\boldsymbol{\epsilon}_\theta$:
\begin{align}
    \tilde{\mathbf{\mu}}_{\theta,t}(\mathbf{x}_t) &= \frac{1}{\sqrt{\alpha_t}} (\mathbf{x}_t - \frac{1 - \alpha_t}{\sqrt{1 - \bar{\alpha}_t}} \boldsymbol{\epsilon}_{\theta}) \\
    \tilde \beta_t &= \frac{1 - \bar{\alpha}_{t-1}}{1 - \bar{\alpha}_t} \cdot \beta_t
\end{align}
%     \tilde{\mathbf{\mu}}_{\theta,t}(\mathbf{x}_t) = \frac{1}{\sqrt{\alpha_t}} (\mathbf{x}_t - \frac{1 - \alpha_t}{\sqrt{1 - \bar{\alpha}_t}} \boldsymbol{\epsilon}_{\theta})
% \end{equation}
% \begin{equation}
% \tilde \beta_t = \frac{1 - \bar{\alpha}_{t-1}}{1 - \bar{\alpha}_t} \cdot \beta_t
% \end{equation}
where $\alpha_t = 1 - \beta_t$, $\bar \alpha_t = \prod_{i=1}^t \alpha_i$.
During training, the model is optimized by the loss proposed by Ho et al.~\cite{ho2020denoising}:
\begin{equation}
    \label{equ:simple}
    \mathcal{L}_{\text{simple}}^t (\theta) = \mathbb{E}_{\mathbf{x}_0,\mathbf{\epsilon}}\left[
    \lVert \mathbf{\epsilon} - \boldsymbol{\epsilon}_\theta (\alpha_t \mathbf{x}_0 + \beta_t \epsilon) \rVert_2^2
    \right].
\end{equation}

\begin{figure*}[t]
\begin{center}
\includegraphics[width=1\textwidth]{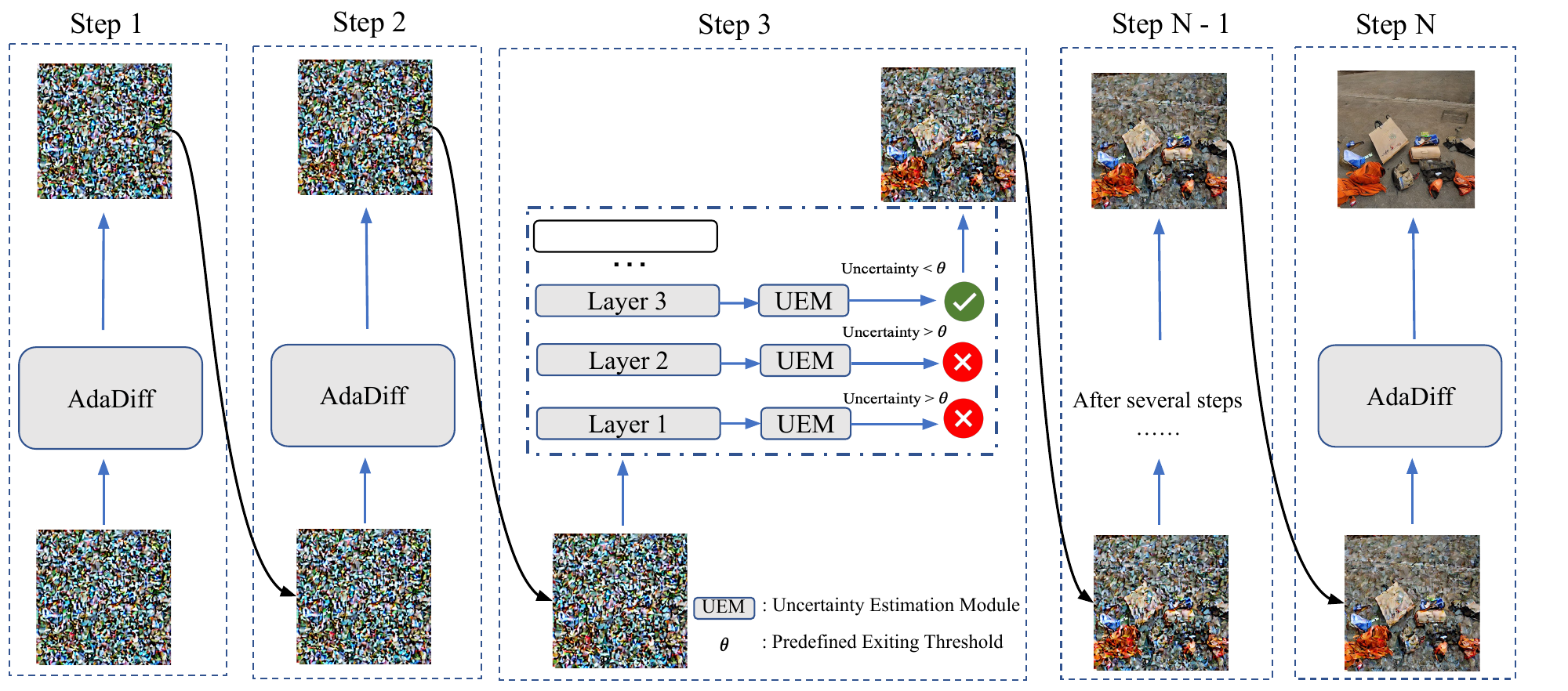}
\end{center}
% \caption{The overview of our proposed AdaDiff. At each step, the output of each intermediate layer is fed into the uncertainty estimation module(UEM) to obtain the uncertainty. Once the uncertainty is lower than the predefined threshold, the following layers will be skipped.} 
\caption{An overview of our proposed AdaDiff method. At each step, the output of each intermediate layer is fed into the Uncertainty Estimation Module (UEM) to quantify the uncertainty. If the estimated uncertainty falls below a predefined threshold, the subsequent layers are skipped, allowing the network to adaptively adjust computation.}
\vspace{-0.5cm}
\label{Fig:Architecture}
\end{figure*}

\subsection{Timestep-Aware Uncertainty Estimation Module}
To detect redundant layers accurately, we propose a timestep-aware uncertainty estimation module attached to each layer to estimate the uncertainty of each hidden state. The whole architecture is illustrated in Figure \ref{Fig:Architecture}. The basic assumption of early exiting is that the input samples of the test set are divided into easy and hard samples. The computation for easy samples terminates once some conditions are satisfied. Most early exiting methods attach a classifier following each intermediate layer to decide whether the network early exits or not. During inference, the output representations of each layer are fed into the classifier to obtain the confidence or entropy which is utilized as a proxy to represent the difficulty of samples and make early exiting decisions. We denote the backbone network consisting of $N$ layers, $L_i$ as the output representation of $i$-th layer, $i \in [1, N]$, and $CLS$ as the classifier. 
\begin{equation}
    C_{cls} =  \max( \frac{1}{1 + e^{-CLS(L_i)}} ) \ or \ E_{cls} = \min (\sum_{c \in C} CLS(L_i)_c \log (CLS(L_i)_c) )
\end{equation}
% \begin{equation}
%     C_{cls} = Softmax(classifier(L_i)) \ or \ E_{cls} = Entropy(classifier(L_i))
% \end{equation}
% \begin{equation}
%     E_{cls} = Entropy(classifier(L_i))
% \end{equation}
If the $C_{cls}$ or $E_{cls}$ meets the predefined requirement such as the confidence is higher or the entropy is lower than the threshold, the computation terminates. Such a strategy works well in classification tasks. However, estimating the noise of each step of diffusion models can be considered a regression task, which makes it challenging to estimate the confidence or entropy of predictions. 

\begin{figure*}[t]
\vspace{-0.2cm}
\begin{center}
\includegraphics[width=0.8\textwidth]{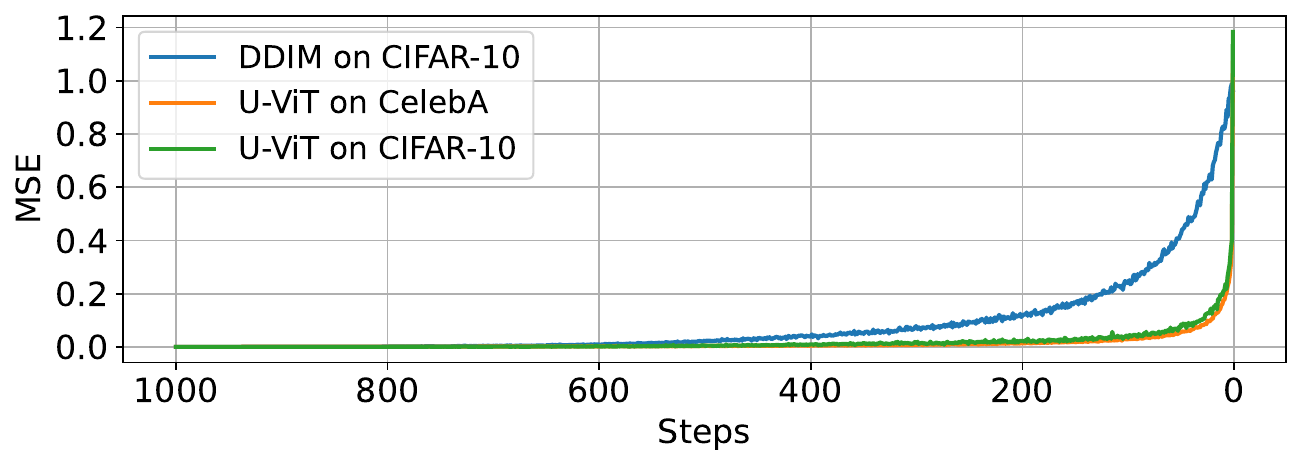}
\end{center}
\vspace{-0.3cm}
% \caption{Average training loss distribution across 1000 steps on CIFAR-10 and CelebA.  The loss difference at different sampling steps demonstrates that the difficulty of denoise is different at each step.}
%\caption{Average MSE across 1000 steps on CIFAR-10 and CelebA.  The loss difference at different sampling steps demonstrates that the difficulty of denoise is different at each step.}
\caption{Average MSE loss across 1,000 denoising steps on the CIFAR-10 and CelebA datasets, where 1000 step is the first generation step and 0 is the final generation step. The varying loss values at different denoising steps indicate that the difficulty of denoising varies throughout the generation process.}
\vspace{-0.3cm}
\label{Fig:training_loss}
\end{figure*}

We consider each generation step separately and assume that the difficulty of the input varies for different generation steps. \cite{hang2023efficient} regards diffusion as multi-task learning and demonstrates that early-generation steps only require simple reconstructions of the input in order to achieve lower denoising loss.
Specifically, we investigate the training loss distribution of each step which can reflect the difficulties of fitting to ground truth with different backbones and datasets, as shown in Figure~\ref{Fig:training_loss}. There is a clear distinction between the difficulty of different sampling steps, which demonstrates that the input of shallow and deep steps can be regarded as easy and hard samples.
Inspired by \cite{xin2021berxit}, we propose a lightweight timestep-aware uncertainty estimation module (UEM) to estimate the uncertainty of each prediction in diffusion models. Specifically, UEM is comprised of a fully-connected layer to estimate the uncertainty $u_i$ of each prediction:
\begin{equation}
    u_{i, t} = f(\mathbf{w_t}^T[L_{i, t},\  timesteps] + b_t)
\end{equation}
where $\mathbf{w_t}, b_t, f, timesteps$ are the weight matrix, weight bias, activation function, and timestep embeddings. We choose sigmoid as an activation function since we hope the output range is $[0, 1]$. If the backbone is based on Transformer that requires tokenization, the output should be unpatched to generate uncertainty maps. The parameters are not shared between different uncertainty estimation modules since the module is able to learn the uncertainty with time-series properties for each layer. To introduce the supervision signal to this module, the pseudo uncertainty ground truth $\hat{u}_i$, which is negatively related
to absolute error of the prediction and ground truth is constructed as:
\begin{equation}
    \hat{u}_{i, t} = F(|g_i(L_{i, t}) - \epsilon_t|)
\end{equation}
where $g_i$ is the output layer and $\epsilon$ is the ground truth noise value. $F$ is a function to smooth the output of absolute error and we utilize the $Tanh$ function in our experiments. The loss function of this module is designed as the MSE loss of estimated uncertainty and pseudo uncertainty ground truth:
\begin{equation}
    \mathcal{L}_{u}^t = \sum_i^N \lVert u_{i, t} - \hat{u}_{i, t} \rVert^2
\end{equation}
During inference, with the estimated uncertainty of the output prediction from each layer, the computation is able to make termination decisions once the uncertainty is lower than the predefined threshold. Moreover, since the generation process of diffusion models requires multiple inferences, we apply such an early exiting strategy to the inference each time.

\subsection{Uncertainty-Aware Layer-wise Loss}
To fill the performance gap between the full model and the accelerated model, we propose an uncertainty-ware layer-wise loss. When a model makes an exit decision, it has to predict the output based on the current input and the information it has learned so far. Although it is confident enough for models to make early exiting decisions, there is always information loss without full layers \cite{teerapittayanon2016branchynet}. To address this issue and retain as much information as possible, previous methods\cite{teerapittayanon2016branchynet, schuster2022confident} utilize a simple layer-wise loss for every layer:
\begin{equation}
    \mathcal{L}_{n}^t = \sum_i^{N-1} \lVert g_i(L_{i, t}) - \epsilon \rVert^2
\end{equation}
This loss function can preserve information with fewer layers and works well in many applications that only require inference once. However, such loss function fails in diffusion models, where multi-step inference is required. The main reason 
is that the information loss accumulates as the sampling proceeds. The error accumulation analysis is supplied in the Appendix. It is challenging for existing layer-wise loss to mitigate the accumulation of errors since they treat all timesteps equally. \cite{hang2023efficient} proved that finetuning specific steps benefits the surrounding steps in mitigating error accumulation. Inspired by this, we propose an uncertainty-ware layer-wise loss:
\begin{equation}
    \mathcal{L}_{UAL}^t = \sum_i^{N-1} (1-u_{i, t}) \times \lVert g_i(L_{i, t}) - \epsilon \rVert^2
\end{equation}
where $u_i$ is the uncertainty value estimated in each layer. With such an uncertainty-aware loss function, timesteps and layers with low uncertainty represent the generation steps and layers that potentially contribute to mitigating the accumulation of errors. Higher weights on such timesteps and layers are the key to filling the performance gap. With the benefit of surrounding timesteps, more timesteps and layers tend to obtain higher certainty, which contributes to skipping more layers and achieving higher efficiency.

\subsection{Joint Training Strategy}
We have already discussed the uncertainty estimation loss in our early exiting framework and uncertainty-aware layer-wise loss for information preservation. There is an interdependence between these two loss functions. 
% Therefore, As shown in Table~\ref{tab:ablation}, directly combining these loss functions might cause unstable training, which in turn leads to performance drop and inefficiency. 
In order to balance the effect between uncertainty estimation loss and uncertainty-aware layer-wise loss, we utilize joint
training strategy to optimize different loss functions simultaneously:
\begin{equation}
    L_{all} = \mathcal{L}_{\text{simple}}^t (\theta) + \lambda\mathcal{L}_u^t + \beta\mathcal{L}_{UAL}^t
\end{equation}
where $\lambda$ and $\beta$ are both hyper-parameters to balance simple task loss, uncertainty estimation loss, and uncertainty-aware layer-wise loss. We set $\lambda$ and $\beta$ both to be 1 in our experiments. 

% The proposed asynchronous training strategy involves optimizing different loss functions at different training steps. Specifically, during odd-numbered training steps, the model optimizes the simple loss function and uncertainty estimation loss, while during even-numbered training steps, the model optimizes the simple loss function and uncertainty-aware layer-wise loss. This strategy allows the model to first optimize uncertainty estimation, which can lead to better optimization with the uncertainty-aware layer-wise loss.

% \clearpage
% Utilizing such asynchronous training strategy, the model is able to optimize uncertainty estimation first. With initial uncertainty estimation, the model achieves better optimization with uncertain-aware layer-wise loss, as we will shown in xxxx.

%% file: section/experiments.tex
\section{Experiments}
\subsection{Experimental Setup}
\noindent \textbf{Dataset.} CIFAR-10~\cite{krizhevsky2009learning} and Celeb-A~\cite{liu2015faceattributes} are used to evaluate our methods and other methods on unconditional generation. ImageNet \cite{deng2009imagenet} is a large-scale dataset with 1000 classes, which is used to evaluate class-conditional generation. Text-guided generation is evaluated on MS-COCO \cite{lin2014microsoft} dataset. The resolution of ImageNet and COCO is both 256$\times$256 while they are 32$\times$32 and 64$\times$64 on CIFAR-10 and CelebA, respectively.

\noindent \textbf{Evaluation protocol.} We measure image generation quality utilizing Frechet Inception
Distance (FID). Following \cite{bao2022all}, we compute FID with 50k generated samples on each task and dataset, with the reference generated from \cite{dhariwal2021diffusion}. As for efficiency, we report the number of average layers used in all test sample generation and its reduction ratio compared with baseline. Moreover, we provide the theoretical GFLOPs of each method on different datasets since the running time is unstable as it is influenced by hardware environments such as memory and I/O. The GFLOPs are calculated as the sum GFLOPs of all steps because the generation is composed of multiple steps.

\noindent \textbf{Baselines.} Our method is based on U-ViT \cite{bao2022all}, a transformer-based diffusion model. On ImageNet 256$\times$256, we utilize the U-ViT-Large model with 21 layers while it is the U-ViT-Small model with 13 layers on other datasets. We compare our method with existing early exiting methods, BERTxiT \cite{xin2021berxit} and CALM \cite{schuster2022confident}, and static early exiting method\cite{moon2023early}. BERTxiT utilizes a learning strategy to extend early exiting to BERT models while CALM uses the similarity of adjacent layers and confidence to decide to exit and calibrates local early exits from global constraints. Static early exiting method\cite{moon2023early} uses a predefined and static early exiting strategy to accelerate generation. We also compare our method with another acceleration method via structural pruning\cite{fang2023structural} referred to as S-Pruning. S-Pruning is an accelerating method that applies structural pruning to diffusion models. Since the results reported in \cite{fang2023structural} are based on 100 sampling steps, for a fair comparison, we also report our results with 100 steps. To show the generalizability of our method, we also conduct experiments on CNN-based diffusion models. The results are shown in the supplementary material.

\noindent \textbf{Implementation details.} Following the settings of existing methods, during training, we utilize the AdamW \cite{loshchilov2017decoupled} optimizer with a learning rate of 2e-4 for all datasets. For all tasks and datasets, we initialize our backbone with pre-trained weight from \cite{bao2022all}. The early exiting threshold is chosen from 0.2 to 0.01. Following \cite{bao2022all}, we utilize DPM-Solver with 50 sampling steps in ImageNet and Euler-Maruyama SDE sampler with 1000 sampling steps in CIFAR-10 and Celeb-A datasets. More implementation details are shown in the Appendix.

 \begin{table*}[t]
 \small
 % \addtolength{\tabcolsep}{0.5pt}
 \renewcommand{\arraystretch}{0.8}
 \renewcommand{\thefootnote}{\fnsymbol[\dagger]}
 % \caption{The performance and efficiency comparison between our method and other early exiting methods with the U-ViT-Small model on CIFAR-10, Celeb-A. Our method reduces computation compared with other methods while preserving the best performance. w/o EE: full models without early exiting.} 
 %\caption{Performance and efficiency comparison between our method and other early exiting methods using the U-ViT-Small model on the CIFAR-10 and CelebA datasets. Our proposed approach achieves the best performance while significantly reducing computational costs compared to other early exiting techniques. By dynamically adapting the network depth based on the complexity of the input data, our method efficiently allocates computational resources, leading to improved efficiency without compromising performance. The "w/o EE" (without Early Exiting) baseline represents the full models without any early exiting mechanism, highlighting the computational savings achieved by our adaptive approach.}
\caption{Performance and efficiency comparison of our proposed method with and without adaptive computation using the U-ViT-Small model for unconditional image generation on the CIFAR-10 and CelebA datasets. The "w/o Adaptive Computation" setting serves as an upper bound for performance without any reduction in computational costs. Our method achieves the best performance in both settings, demonstrating its effectiveness in maintaining high accuracy while enabling computational savings through adaptive depth adjustment. }
 \vspace{-0.3cm}
    \centering
    \begin{tabular*}{12.3cm}{lccccccc}
    \toprule
    \toprule
        \multirow{2}*{Methods} & 
        \multicolumn{3}{c}{CIFAR-10 32$\times$32} & &
        \multicolumn{3}{c}{CelebA 64$\times$64}\\
        \cmidrule{2-4}
        \cmidrule{6-8}
         & FID  & Layers Ratio &GFLOPs & & FID & Layers Ratio &GFLOPs \\
   \midrule
\multicolumn{5}{l}{\textit{Models w/o Adaptive Computation}}  \\
       U-ViT &3.11 & 1 &22.86 &  &2.87 & 1 &23.01   \\
       Ours &\textbf{2.7} & 1 &22.86 &  &\textbf{2.63} & 1 &23.01    \\
       \midrule
    \multicolumn{5}{l}{\textit{Models w/ Adaptive Computation} } \\
    % \cline{2-8}
     % \hline
       BERxiT &20.5 &-20.5\% & 18.18 (-20.4\%) &  &31.01 & -18.5\% &18.79 (-18.3\%)   \\
       CALM &20.0 & -20.5\% &18.18 (-20.4\%) & &25.3 & -20.5\% &18.33 (-20.3\%)    \\
       Static EE &- & - &- & &5.04 &-32.09\% &-    \\
    % \cline{2-8}
       Ours &\textbf{3.7} & \textbf{-47.7\%} &\textbf{11.97 (-47.6\%)}  & &\textbf{3.9} & \textbf{-46.2\%} &\textbf{12.48 (-45.7\%)}    \\
     \bottomrule
     \bottomrule
    \end{tabular*}
    \label{tab:unconditional}
    \vspace{-5mm}
\end{table*}

%  \begin{table*}[t]
%  % \addtolength{\tabcolsep}{0.5pt}
%  \renewcommand{\arraystretch}{1}
%  \renewcommand{\thefootnote}{\fnsymbol[\dagger]}
%  \caption{The performance and efficiency comparison between our method and S-Pruning on CIFAR-10, Celeb-A. Our method reduces computation compared with other methods while preserving well performance. All methods utilize 100 sampling steps.} \vspace{-0.3cm}
%     \centering
%     \begin{tabular*}{6.8cm}{lccccccc}
%     \toprule
%     \toprule
%         \multirow{2}*{Methods} & 
%         \multicolumn{2}{c}{CIFAR-10 32$\times$32} & &
%         \multicolumn{2}{c}{CelebA 64$\times$64}\\
%         \cmidrule{2-3}
%         \cmidrule{5-6}
%          & FID  &Layer Ratio & & FID  &Layer Ratio \\
%     \midrule
%        S-Pruning  &5.29 & -44.2\%  & &6.24 & -44.3\%     \\
%        Ours  &4.12 & -47.7\%  & &4.67 & -46.2\%     \\
%      \bottomrule
%      \bottomrule
%     \end{tabular*}
%     \label{tab:unconditional}
% \end{table*}

\begin{wraptable}{r}{0.6\textwidth}
 \renewcommand{\arraystretch}{0.8}
 \vspace{-12mm}
    \caption{
    The performance and efficiency comparison between our method and S-Pruning on CIFAR-10, Celeb-A. Our method achieves better FID with a similar layer reduction ratio as S-Pruning. All methods utilize DDIM 100 sampling steps.}
    
\begin{tabular}{ccccccccp{2cm}|}
\toprule
\toprule
\multirow{2}*{Methods} &
 \multicolumn{2}{c}{CIFAR-10 32$\times$32}  & &
 \multicolumn{2}{c}{CelebA 64$\times$64} \\
 \cmidrule{2-3}
 \cmidrule{5-6}
  & FID & Layer Ratio & & FID & Layer Ratio \\
\hline
S-Pruning  &5.29 & -44.2\%  & &6.24 & -44.3\%     \\
       Ours  &4.12 & -47.7\%  & &4.67 & -46.2\%     \\
     \bottomrule
     \bottomrule
\end{tabular}
    \label{tab:unconditional_spruning}

    \vspace{-8mm}
\end{wraptable}

% \begin{table}[t]
%     \centering
%     \caption{The performance and efficiency comparison between our method and other early exiting methods with the large model on MS-COCO dataset. w/o EE: without early exiting or the exiting threshold is 0.}
%     \begin{tabular*}{10.5cm}{l|c|ccc}
%     \toprule
%     \toprule
%         \multirow{2}*{Model} &\multirow{2}*{Methods} & \multicolumn{3}{c}{MS-COCO} \\
%         & & FID  & Layers Ratio &GFLOPs \\
%     \hline
%        &Baseline &5.95 & 1 &25.04\\
%     \cline{2-5}
%        &BERxiT &18.9 & -10.0\%  &22.53 (-10.02\%)\\
%       U-ViT-Small &CALM &17.0 & -15.4\% &21.18 (-15.4\%)  \\
%     \cline{2-5}
%         % &Ours w/o EE &\textbf{6.22} & -0\%   &25.04\\
%        &Ours &\textbf{7.40} & \textbf{-43.6\%} &\textbf{14.12 (43.6\%)}  \\
%     \bottomrule
%     \bottomrule
%     \end{tabular*}
%     \label{tab:coco}
% \end{table}

\subsection{Unconditional and Class-Conditional Generation}
\label{sec:unconditional}
We conduct unconditional generation experiments on CIFAR-10, Celeb-A, and class-conditional generation on ImageNet $256 \times 256$. As shown in Figure~\ref{tab:unconditional}, our method achieves the best efficiency with minimal performance loss compared with other early exiting methods on CIFAR-10 and CelebA. More concretely, our method can reduce 47.7\% layers on average and achieve 3.7 in FID while other early exiting methods such as BERTxiT and CALM only obtain layer reduction in around 20\% and intense performance drop to 20 in FID on CIFAR-10 dataset. Moreover, on Celeb-A, the performance drop of BERTxiT and CALM is more severe with around 31 and 25 in FID and less than 20\% layer number reduction while our method gains 3.9 in FID and uses 46.2\% less layers. Surprisingly, the full model that is trained via our AdaDiff obtains 2.7 and 2.63 FID on CIFAR-10 and CelebA, which is better than U-ViT. We also compare our method with another acceleration method such as S-Pruning in Figure~\ref{tab:unconditional_spruning}.  Our method achieves better FID with a similar layer reduction ratio as S-Pruning. On ImageNet, as shown in Figure~\ref{tab:imagenet_coco}, the FID of our method is 4.5 while the layer number is reduced to around 45.2\%. In contrast, BERTixT and CALM incur severe performance drops to 23.5 and 21.4 respectively with less than 20\% efficiency. The main reason is that our method helps the model to learn to use fewer layers to preserve more information and therefore achieve high efficiency with minimal performance loss.

\begin{table}[t]
 \renewcommand{\arraystretch}{0.8}
% \caption{The performance and efficiency comparison between our method and other early exiting methods on ImageNet and COCO dataset. }
\caption{Performance and efficiency comparison of our proposed method with and without adaptive computation using the U-ViT-Small model for conditional image generation on the ImageNet and COCO datasets. The "w/o Adaptive Computation" setting serves as an upper bound for performance without any reduction in computational costs. Our method achieves the best performance in both settings, demonstrating its effectiveness in maintaining high accuracy while enabling computational savings through adaptive depth adjustment. }
\vspace{-0.3cm}
    \centering
    \begin{tabular*}{12.3cm}{lccccccc}
    \toprule
    \toprule
         \multirow{2}*{Methods} 
         & \multicolumn{3}{c}{ImageNet 256$\times$256} &
         & \multicolumn{3}{c}{MS-COCO 256$\times$256} \\
         \cmidrule{2-4}
        \cmidrule{6-8}
         & FID  & Layers Ratio &GFLOPs & & FID  & Layers Ratio &GFLOPs \\
    \midrule
    \multicolumn{5}{l}{\textit{Models w/o Adaptive Computation}}  \\
       U-ViT &3.4 & 1 &142.20 & &5.95 & 1 &25.04\\
       Ours &3.61 & 1 &142.20 &  &6.12 & 1 &25.04 \\
       \midrule
    % \cline{2-5}
    \multicolumn{5}{l}{\textit{Models w/ Adaptive Computation}}  \\
       BERxiT &23.5 & -17.8\% & 116.89 (-17.7\%) & &18.9 & -10.0\%  &22.53 (-10.02\%) \\
       CALM &21.4 & -19.3\% &114.76 (-19.2\%) & &17.0 & -15.4\% &21.18 (-15.4\%)  \\
        % &Ours w/o EE &\textbf{3.5} & 0\% &142.20  \\
       Ours &\textbf{4.5} & \textbf{-45.2\%} &\textbf{77.88 (-45.2\%)} &  &\textbf{7.40} & \textbf{-43.6\%} &\textbf{14.12 (43.6\%)}  \\
    \bottomrule
    \bottomrule
    \end{tabular*}
    \label{tab:imagenet_coco}
    \vspace{-4mm}
\end{table}

\begin{figure*}[t]
    \centering
    \includegraphics[width=0.8\textwidth]{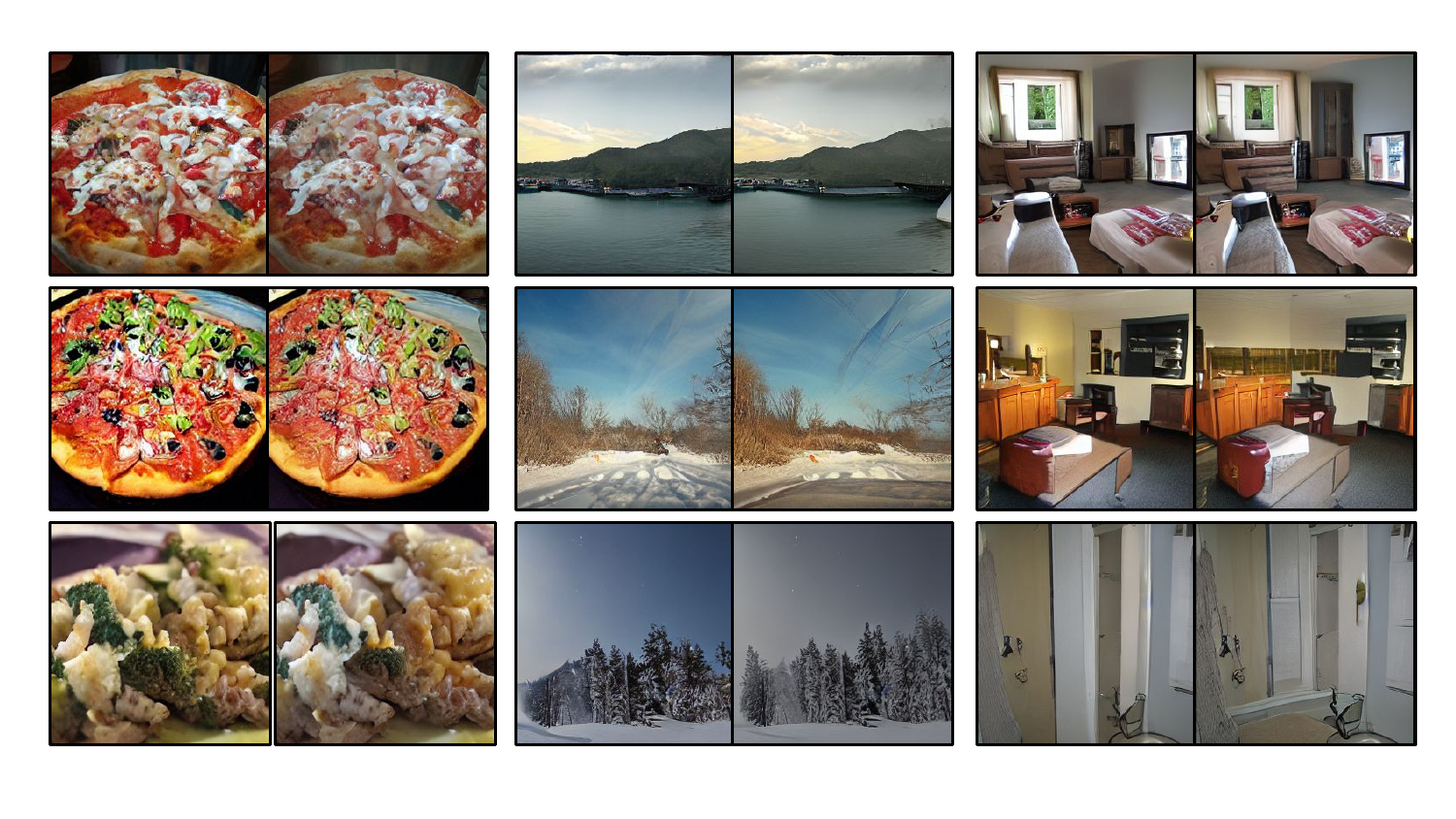}
    \vspace{-6mm}
    \caption{Generation samples comparison between the model w/o adaptive computation (left) and model w/ adaptive computation (right) on COCO.}
    \vspace{-0.5cm}
    \label{fig:sample_coco}
\end{figure*}

\subsection{Text-guided Generation}
\label{sec:text}
In this section, we discuss the text-guided generation on the MS-COCO dataset. Like on ImageNet, we utilize DPM-Solver with 50 sampling steps on MS-COCO. The main quantitative results are shown in Table ~\ref{tab:imagenet_coco}. Our method achieves the best performance with high efficiency compared with other methods on MS-COCO. Specifically, AdaDiff obtains 7.4 FID with 43.6\% computation reduction. In comparison, the FID of BERTxiT and CALM is 23.5 and 21.4 with 17.8\% and 19.3\% layer number reduction, respectively. The GFLOPs results are consistent with the layer reduction ratio we report in the table. Also, we illustrate the visual results on COCO shown in Figure~\ref{fig:sample_coco}. From the nine image groups, we can observe that the early-exited model can generate nearly the same high-fidelity image as full models without distortions. In some cases, our method can generate images with more 
realistic details, such as the last image group in the middle. In all, our method can generate samples that resemble the full model.

% Moreover, we provide several generation samples comparison between our method and existing early exiting methods. As shown in Figure~\ref{Fig:sample}, the generation samples of BERTxiT and CALM consist of distortions in images such as bus and people in the images while our results include fewer distortions with more fidelity. The main reason is that BERTxiT and CALM lose much information during denoising, which is unable to estimate accurate noise distribution. In contrast, our method is able to preserve as much information with less number of network layers.

% \begin{figure}[t]
% \begin{center}
% \includegraphics[width=1\textwidth]{sections/Figures/generation.pdf}
% \end{center}
% \caption{Text-guided generation sample from MS-COCO.}
% \label{Fig:sample}
% \end{figure}
\vspace{-3mm}
\subsection{Experimental Analysis of AdaDiff}
\vspace{-2mm}

% \noindent \textbf{Generation Samples Comparison.} We show the generation sampling qualitative comparison with the same prompt on COCO datasets in Fig~\ref{fig:sample_coco}. More generation samples can be found in the appendix. From the nine image groups, we can observe that the early-exited model can generate nearly the same high-fidelity image as full models without distortions. In some cases, our method can generate images with more 
% realistic details, such as the last image group in the middle. In all, our method can generate samples that resemble the full model.

\begin{figure*}[t]
\begin{center}
\subfloat[CIFAR-10]{\includegraphics[width=0.32\textwidth]{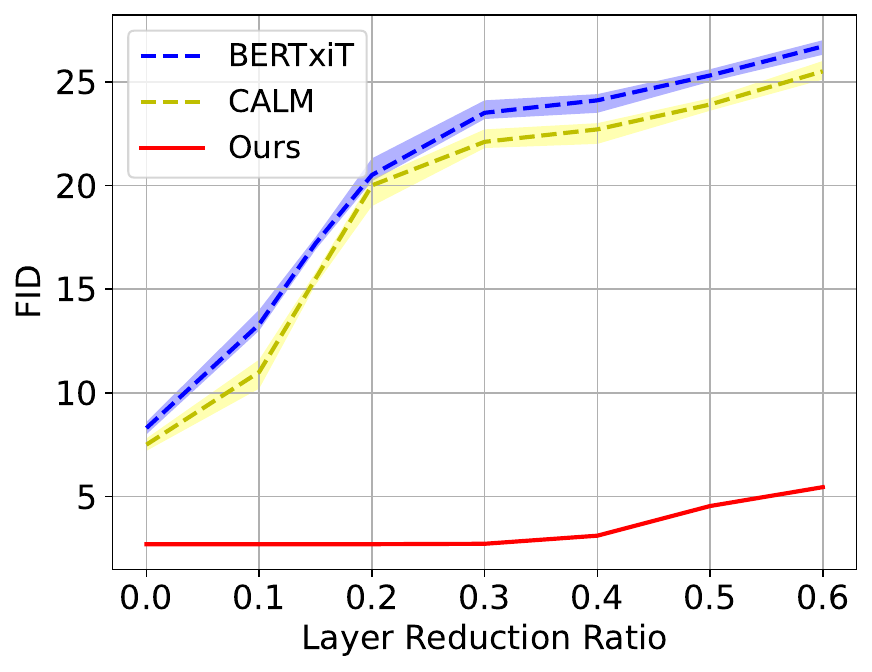}}
\subfloat[Celeb-A]{\includegraphics[width=0.32\textwidth]{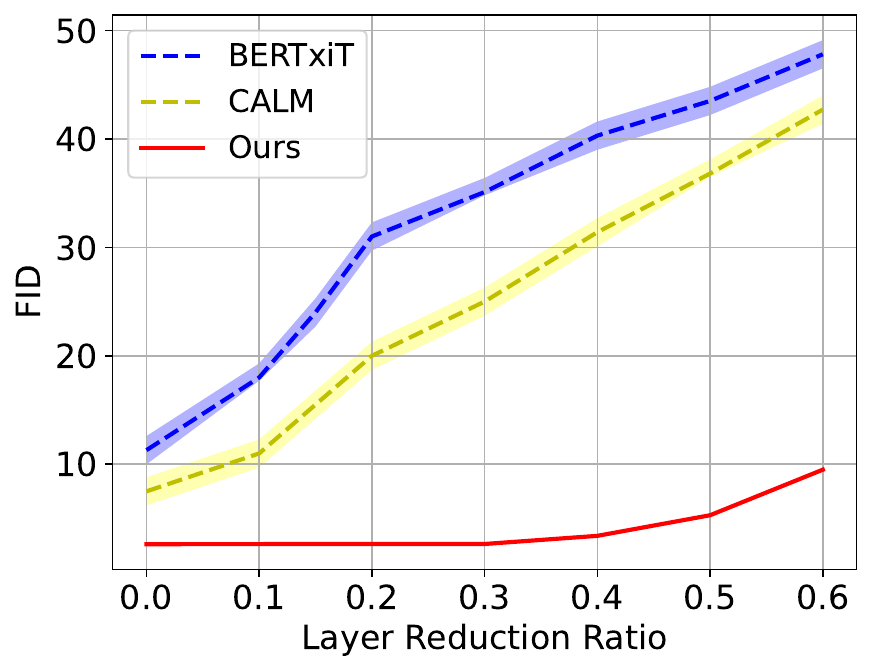}}
\subfloat[MS-COCO]{\includegraphics[width=0.32\textwidth]{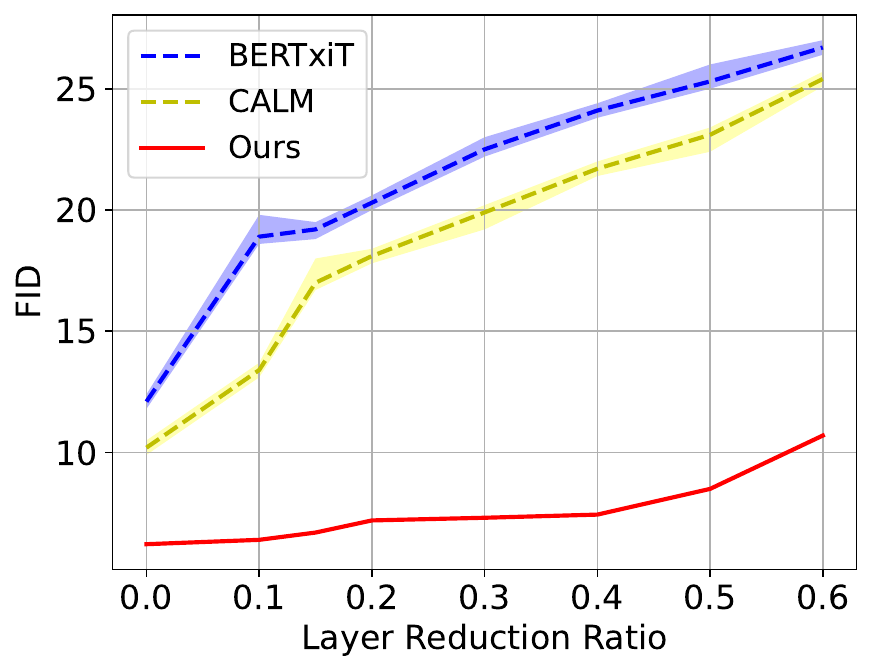}}
\end{center}
\vspace{-0.2cm}
% \caption{Performance-Efficiency trade-off Curve on CIFAR-10, CelebA, COCO. Our method achieves the best performance and efficiency trade-off compared with other methods. Moreover, our method with a 50\% acceleration ratio obtains similar performance as other methods without acceleration. }
\caption{Performance-Efficiency trade-off curve on the CIFAR-10, CelebA, and COCO datasets. The trade-off curve demonstrates that our method maintains much lower FID scores, indicating higher quality generated images, while having the same layer reduction ratio as other adaptive computational methods.}
\vspace{-0.5cm}
\label{Fig:trade-off}
\end{figure*}
\noindent \textbf{Performance and Efficiency Trade-off.} According to the discussion in Sec~\ref{sec:unconditional} and ~\ref{sec:text}, our AdaDiff has achieved the best efficiency with minimal performance drop compared with other early exiting methods on all datasets. In this section, we discuss the performance and efficiency trade-off of our methods and other frameworks. We report the performance and efficiency trade-off curve of AdaDiff, BERTxiT, and CALM on CIFAR-10, Celeb-A, and MS-COCO as shown in Figure~\ref{Fig:trade-off}. First of all, without doing early exiting, our method obtains the best FID value compared with BERTxiT and CALM. Moreover, BERTxiT and CALM tend to cause large performance drops with the increase of layer reduction ratio. In contrast, the trade-off curve of our AdaDiff is flat, which means our method can stay stable when more layers are skipped. Interestingly, at the highest
efficiency point, our method achieves similar performance to other methods such as BERTxiT and CALM at the lowest computation reduction point on CIFAR-10, which demonstrates the effectiveness of our method.

% \noindent \textbf{Statistics of Exited Layers.} We provide more experimental analysis about our proposed dynamic early exiting strategy. We report the average exited layer ratio of our model with different thresholds on the CIFAR-10 dataset, as shown in Figure~\ref{Fig:layer}. In the left figure, the experiment is conducted with a relatively low uncertainty threshold. Most computations are fully operated while there are still some parts of computation that are early stopped. In contrast, with a higher uncertainty threshold, most of the computations are terminated at the beginning which means some easy samples are allocated with less computation. This shows that with threshold control, our estimated uncertainty achieves a good trade-off between quality and efficiency.
% \begin{figure*}[t]
% \begin{center}
% \subfloat[]{\includegraphics[width=0.3\textwidth]{fig/Layer_small.pdf}}
% \subfloat[]{\includegraphics[width=0.3\textwidth]{fig/Layer_large.pdf}}
% \end{center}
% \caption{Statistics of exited layers on CIFAR-10. \textcolor{red}{[DK: please use "vspace" command for each table and figure to further reduce the blank space. please contact DK if not sure about how to use the command]}, \textcolor{blue}{[DK: add 1-2 sentences simply describing the main observation from this figure and its indication/conclusion]}}
% \label{Fig:layer}
% \end{figure*}

\noindent \textbf{Performance comparison between our models with small models trained from scratch.} To show the effectiveness of our method on top of the baseline model, we report the results comparison of our method with a small-size model that is trained from scratch. All training settings of the small model stay unchanged compared with baseline models except for layer number. The small model is comprised of 7 layers, which have similar GFLOPs as our model. Also, we choose the best evaluation epoch for a fair comparison. The results are shown in Table \ref{tab:small_model}. Our method achieves better performance than the small model on CIFAR-10 and CelebA, which indicates the effectiveness of our method in accelerating the inference of diffusion compared with purely training a small model.

\begin{wraptable}{r}{0.6\textwidth}
\renewcommand{\arraystretch}{0.8}
 \vspace{-12mm}
     % \caption{{Results comparison of small-size models, static exiting models and our method on CIFAR-10 and CelebA. The superior performance of our model demonstrates the effectiveness of our proposed AdaDiff.}} 
     \caption{Comparative analysis of our AdaDiff method against small-size models and static early exiting approaches on the CIFAR-10 and CelebA datasets. We evaluate the performance of these methods under similar computational budgets and performance levels. Small-size models with comparable computational costs and static early exiting models with varying fixed layer reduction ratios serve as baselines, matching either the computational budget or the performance level of AdaDiff.}
    \begin{tabular}{ccccc}
    \toprule
    \toprule
         \multirow{2}*{Methods} & \multicolumn{2}{c}{CIFAR-10} &\multicolumn{2}{c}{CelebA} \\
         \cmidrule{2-3}
         \cmidrule{4-5}
         & FID  & GFLOPs  & FID  & GFLOPs  \\
    \midrule
       Small model &6.68 & 12.8  &4.58 & 13.0  \\
       \midrule
        Exit at 11th layer &4.1 &19.19 &4.5 &19.47 \\
        % Exit at 9th layer  &6.2 &15.70 &6.9 &15.93 \\
        Exit at 7th layer  &8.3 &12.21 &8.8 & \textbf{12.39} \\
        \midrule
       Ours & \textbf{3.7} & \textbf{11.97}  & \textbf{3.9}  &12.48  \\
    \bottomrule
    \bottomrule
    \end{tabular}
     \vspace{-7mm}
  
    \label{tab:small_model}
\end{wraptable}
\noindent \textbf{Results Comparison with Static Exiting Models.} 
To show the effectiveness of our dynamic early exiting method, we compare our method with the static exiting method, namely exits at a certain layer for all samples.
The static model is trained with our proposed UEM and uncertainty-aware layer-wise loss. All training settings are the same as our model.
The reported results are shown in Table \ref{tab:small_model}. In the table, the model that exits at the 7th layer shares similar GFLOPs as our dynamic method. Our method achieves 3.7 and 3.9 FID on CIFAR-10 and CelebA while the static model only obtains 8.3 and 8.8 FID on each dataset, which demonstrates the effectiveness of our adaptive strategy.

\begin{figure}[t]
\vspace{-2mm}
\begin{center}
\captionsetup[subfloat]{labelsep=none, format=plain, labelformat=empty}
\subfloat[Result]{\includegraphics[width=0.15\textwidth]{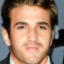}}
\hspace{0.1cm}
\subfloat[100 steps]{\includegraphics[width=0.15\textwidth]{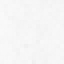}}
\hspace{0.1cm}
\subfloat[300 steps]{\includegraphics[width=0.15\textwidth]{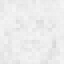}}
\hspace{0.1cm}
\subfloat[500 steps]{\includegraphics[width=0.15\textwidth]{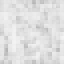}}
\hspace{0.1cm}
\subfloat[700 steps]{\includegraphics[width=0.15\textwidth]{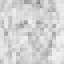}}
\hspace{0.1cm}
\subfloat[900 steps]{\includegraphics[width=0.15\textwidth]{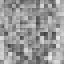}}
\end{center}
\vspace{-0.2cm}
\caption{The visualization of uncertainty map on different sampling steps of our AdaDiff. The figure shows that the uncertainty will increase during generation, which aligns with the observation that the difficulty of different steps varies.} 
\vspace{-0.5cm}
\label{Fig:uncertainty_map}
\end{figure}

\noindent \textbf{Uncertainty Map Visualization.} In order to demonstrate the effectiveness of our proposed uncertainty-aware layer-wise loss, we provide a group of images that visualize the uncertainty map at different sampling steps as shown in Figure~\ref{Fig:uncertainty_map}. At the beginning of the reverse denoising process, the uncertainty estimated by the UEM module is insignificant, which means shallow layers can generate similar results as deep layers at the early generation stage. However, as the input of the backbone includes less noise, the uncertainty value tends to increase. The larger the number of steps sampled, the higher the uncertainty of generating similar results as deep networks using early exited networks. The reason is that at the late stages of generation, more fine-grained noise estimation is required, which can be treated as hard examples for early exiting networks, resulting in more computation resource allocation, which aligns with the observation in Figure~\ref{Fig:training_loss}. Moreover, the uncertainty map shows the uncertainty structures such as the face structure in the images, demonstrating the accuracy of uncertainty estimation.

\begin{wraptable}{r}{0.55\textwidth}
\vspace{-4mm}
 \renewcommand{\arraystretch}{0.9}
\caption{Results with DPM-solver in unconditional image generation on CelebA. Our method can further improve efficiency by integrating with other acceleration methods.}
    \centering 
    \begin{tabular}{ccc}
    \toprule
    \toprule
        Models &FID &Layers Ratio \\
    \hline
         AdaDiff + DPM (Full) & 1.8 & 1 \\
         U-ViT+ DPM &3.32 & 1 \\
         AdaDiff + DPM  w/ EE & 3.9 & -42.3 \% \\
    \bottomrule
    \bottomrule
    \end{tabular}
    \label{tab:ablation}
    \vspace{-8mm}
\end{wraptable}

\noindent \textbf{Different Sampling Strategies.}
We also conduct several experiments with DPM-Solver in 50 steps to show our model can combine with other acceleration methods. Surprisingly, the model achieves even better results without early exiting compared with our basic model, demonstrating the ability to be well combined with other acceleration methods. With early exiting, AdaDiff obtains similar performance as the basic model. More discussion can be found in the appendix.

% \begin{table}[ht]
% \renewcommand{\arraystretch}{1}
% \caption{Ablation study in unconditional image generation on CelebA. The results in the first four lines demonstrate the effectiveness of our proposed joint training strategy and uncertainty-aware layer-wise loss. The results in the last three lines show that our can adapt to other acceleration methods.}
%     \centering 
%     \begin{tabular}{l|c|c}
%     \toprule
%     \toprule
%         Models &FID &Layers Ratio \\
%     \hline
%          U-ViT+ DPM &3.32 & - \\
%          AdaDiff + DPM w/o EE & 1.8 & - \\
%          AdaDiff + DPM & 3.9 & -42.3 \% \\
%     \bottomrule
%     \bottomrule
%     \end{tabular}
%     \label{tab:ablation}
% \end{table}

\vspace{-4mm}
\subsection{Ablation Study}
\vspace{-1mm}
% \begin{table}[ht]
% \renewcommand{\arraystretch}{1}
% \caption{Ablation study in unconditional image generation on CelebA. The results in the first four lines demonstrate the effectiveness of our proposed joint training strategy and uncertainty-aware layer-wise loss. The results in the last three lines show that our can adapt to other acceleration methods.}
%     \centering 
%     \begin{tabular}{l|c|c}
%     \toprule
%     \toprule
%         Models &FID &Layers Ratio \\
%     \hline
%          AdaDiff w/o EE &2.63 &-  \\
%          AdaDiff w/o UA-Loss &14.5 &-20\% \\
%          AdaDiff + Parameter Sharing &5.7 &-41.7\% \\
%          AdaDiff &3.9 &-46.2\%  \\
%     \bottomrule
%     \bottomrule
%     \end{tabular}
%     \label{tab:ablation}
% \end{table}
\noindent \textbf{AdaDiff without UA-Loss.}
To analyze how our proposed uncertainty-aware \begin{wraptable}{r}{0.5\textwidth}
\vspace{-12mm}
% \caption{Ablation study in unconditional image generation on CelebA. The results in the first line demonstrate the effectiveness of our proposed uncertainty-aware layer-wise loss. The results in the second line show that without parameter sharing, the model can generate better uncertainty estimation and thus better performance.}
\caption{Ablation study on the effectiveness of key components in our AdaDiff method for unconditional image generation on the CelebA dataset.}
    \begin{tabular}{ccc}
    \toprule
    \toprule
        Models &FID &Layers Ratio \\
    \hline
     AdaDiff &3.9 &-46.2\%  \\
         % AdaDiff w/o EE &2.63 &-  \\
         AdaDiff w/o UA-Loss &14.5 &-20\% \\
         AdaDiff + Sharing &5.7 &-41.7\% \\
        
    \bottomrule
    \bottomrule
    \end{tabular}
    \vspace{-8mm}
    \label{tab:ablation}
    \vspace{-1mm}
\end{wraptable} 
layer-wise loss affects the performance, we conduct an ablation study that excludes uncertainty-aware layer-wise loss. 
As shown in Table~\ref{tab:ablation}, the results on the first line show that the performance of the model without UA-Loss degrades to 14.5 with only 20\% layer reduction, reflecting that our UA-loss can fill the performance gap between early-exited models and full models.

\noindent \textbf{Parameter Sharing.}
We compare the model with parameter sharing and our models. The performance shown on the second line in Table~\ref{tab:ablation}
is 5.7 FID with 41.7\% layer drop. We believe parameter sharing will be harmful to the uncertainty estimation of diffusion models since the large sampling steps and number of layers make it hard for a simple linear layer to learn fine-grained uncertainty.

 % We apply our proposed joint training strategy on BERTxiT and CALM. Moreover, we test the performance and efficiency of DeeDiff without uncertainty-aware layer-wise loss and joint training strategy, respectively. The qualitative results are shown in Figure~\ref{tab:ablation}. From the first and second groups of experiments, with our proposed joint training strategy, BERTxiT and CALM both obtain benefits in performance and efficiency compared with the original results. More specifically, BERTxiT and CALM originally gain 20.5 and 20.0 FID with both 20.5\% layer reduction while achieving 17.3 and 15.7 with 22\% and 25\% layer reduction respectively with a joint training strategy. This demonstrates the effectiveness of the joint training strategy, which is able to balance the layer-wise loss and uncertainty estimation. Besides, the performance and efficiency of DeeDiff without uncertainty-aware layer-wise loss drop intensely. Concretely, such a model only gains 14.5 FID with 30\% layer reduction, which reversely proves the benefits obtained from our proposed loss. 

%% file: section/conclusion.tex
\vspace{-2mm}
\section{Conclusion \& Discussion}
\vspace{-2mm}
In this work, we propose a dynamic early exiting method called AdaDiff to accelerate diffusion model generation. The timestep-aware uncertainty estimation module(UEM) is aimed at estimating the uncertainty of the output. Furthermore, our uncertainty-aware layer-wise loss concentrates on filling the performance gap. Our method achieves SoTA performance compared with other early exiting methods. Specifically, our method obtains 3.7, 3.9, 4.5 and 7.4 FID with more than 40\% acceleration ratio on CIFAR-10 CelebA, ImageNet and COCO.

However, there are limitations to our methods. First, although the performance with fixed efficiency of our method is the best among other methods, AdaDiff still obtains high FID when the efficiency increases more than 60\%. The potential reason behind the results could be the limit of expression for exited models.  Second, our AdaDiff only considers the height of the generation process. The width of the generation process, namely the adaptive sampling steps is unexplored and left to future works.